%% file: main.tex
\documentclass{article}

\usepackage[preprint]{neurips_2023}

\usepackage[utf8]{inputenc} 
\usepackage[T1]{fontenc}    
\usepackage[colorlinks=true,linkcolor=blue,citecolor=blue,urlcolor=black]{hyperref}       
\usepackage{url}            
\usepackage{booktabs}       
\usepackage{amsfonts}       
\usepackage{nicefrac}       
\usepackage{microtype}      
\usepackage{xcolor}         
\usepackage{caption} 

\usepackage{algorithm}
\usepackage{algorithmicx}
\usepackage{algpseudocode}

\usepackage{amsmath}
\usepackage{amssymb}
\usepackage{mathtools}
\usepackage{amsthm}
\input{math_commands.tex}

\usepackage[capitalise]{cleveref} 

\newcommand{\bl}[1]{\textcolor{blue}{#1}}

\newcommand{\modelnameA}{{\sc Odin}}

\title{\modelnameA{}: Disentangled Reward \\Mitigates Hacking in RLHF}

\author{ Lichang Chen$^{*}$\textsuperscript{$\ddagger$} \and \textbf{Chen Zhu}$^{*}$\textsuperscript{$\dagger$} \and \textbf{Davit Soselia}\textsuperscript{$\ddagger$} \and \textbf{Jiuhai Chen}\textsuperscript{$\ddagger$} \and \textbf{Tianyi Zhou}\textsuperscript{$\ddagger$} \and \hspace{-1em}\textbf{Tom Goldstein}\textsuperscript{$\ddagger$} \and \textbf{Heng Huang}\textsuperscript{$\ddagger$} \and \textbf{Mohammad Shoeybi}\textsuperscript{$\dagger$} \and \textbf{Bryan Catanzaro}\textsuperscript{$\dagger$}}

\begin{document}

\maketitle
\def\thefootnote{\textsuperscript{$\dagger$}}\footnotetext{ NVIDIA \textsuperscript{$\ddagger$} University of Maryland, College Park
}\def\thefootnote{\arabic{footnote}}
\def\thefootnote{\textbf{*}}\footnotetext{Equal contribution, order is random. 
Correspondence to Lichang Chen <bobchen@cs.umd.edu>, Chen Zhu <chzhu@nvidia.com>. \\
}\def\thefootnote{\arabic{footnote}}

\begin{abstract}
In this work, we study the issue of reward hacking on the response length, a challenge emerging in Reinforcement Learning from Human Feedback (RLHF) on LLMs. A well-formatted, verbose but less helpful response from the LLMs can often deceive LLMs or even human evaluators to achieve high scores.  The same issue also holds for some reward models in RL. To address the challenges in both training and evaluation, we establish a more reliable evaluation protocol for comparing different training configurations, which inspects the trade-off between LLM evaluation score and response length obtained by varying training hyperparameters. Based on this evaluation, we conduct large-scale studies, where the results shed insights into the efficacy of hyperparameters and tricks used in RL on mitigating length bias. We further propose to improve the reward model by jointly training two linear heads on shared feature representations to predict the rewards, one trained to correlate with length, and the other trained to decorrelate with length and therefore focus more on the actual content. We then discard the length head in RL to prevent reward hacking on length. Experiments demonstrate that our approach almost eliminates the reward correlation with length, and improves the obtained policy by a significant margin.

\end{abstract}

\input{introduction}

\input{context_challenges}

\input{method}
\input{experiment}

\input{related_works}
\input{conclusion}

\bibliography{main}
\bibliographystyle{abbrvnat}

\clearpage
\appendix
\input{appendix}


\end{document}

%% file: math_commands.tex
\usepackage{amsmath,amsfonts,bm}

\def\real{{\mathbb{R}}}

\def\eqref#1{equation~\ref{#1}}

\def\1{\bm{1}}
\newcommand{\train}{\mathcal{D}}

\def\rmW{{\mathbf{W}}}

\def\vtheta{{\bm{\theta}}}

\def\vw{{\bm{w}}}

\DeclareMathAlphabet{\mathsfit}{\encodingdefault}{\sfdefault}{m}{sl}
\SetMathAlphabet{\mathsfit}{bold}{\encodingdefault}{\sfdefault}{bx}{n}

\def\gB{{\mathcal{B}}}

\def\gD{{\mathcal{D}}}

\def\sD{{\mathbb{D}}}

\newcommand{\E}{\mathbb{E}}
\newcommand{\Ls}{\mathcal{L}}



%% file: introduction.tex
\section{Introduction}
\label{sec: introduction}

Reinforcement Learning from Human Feedback (RLHF) has emerged as a critical technique to elicit the capabilities from pretrained large language models~(LLMs) to generate more helpful, honest, and harmless responses that align with human preferences~\citep{ziegler2019finetuning,askell2021general,instructgpt}, which has led to the success of ChatGPT~\citep{chatgpt} and many other AI systems~\citep{bard,claude,touvron2023llama}.
RLHF trains a reward model~(RM) on human preferences for the responses of given prompts, followed by training the language model to generate responses that maximize the learned reward through reinforcement learning. 
Such a paradigm simplifies human data collection, as acquiring human ratings is easier than collecting demonstrations for supervised fine-tuning. 
Moreover, it has been observed that RLHF has weak-to-strong generalization, where the policy becomes more creative than the supervision it receives~\citep{burns2023weak}.

Despite the promises, one subtle issue of RLHF is reward hacking, or reward model over-optimization, \textit{i.e.}, the policy obtains a high reward but does not fulfill the actual objectives. 
It happens because the RM is not a perfect proxy of human preferences and has limited out-of-distribution~(OOD) generalization, but the policy is a capable LLM that can learn to generate OOD examples to exploit the vulnerabilities of the RM~\citep{hendrycks2021unsolved,rame2024warm}.
More critically, the human preference data can often be biased and inconsistent due to the difficulty and subjectivity of the task itself, flaws in the rating criteria, and the limited quality of raters. 
The most common pattern of reward hacking in practice is verbosity: the language models generate more tokens to make the response appear more detailed or better formatted after RLHF (usually for helpfulness) but the actual quality does not improve~\citep{long-way-to-go,wang2023far}.
This tendency is largely due to a preference among human raters for longer responses, which could be exploited by RM easily and cause the length hacking.
Given the challenges in controlling the quality of human data, it becomes increasingly important and beneficial to study mitigating the impact of spurious features from the reward modeling and algorithmic perspective.

In this paper, we take a step towards mitigating reward hacking by conducting a comprehensive study on the impact of reward models and the RL algorithm on the verbosity and performance of the learned policy.
Considering the challenges in model-based evaluations due to their biases~\citep{zeng2023evaluating}, \textit{e.g.}, open-sourced LLMs climb up on Alpaca-Eval~\citep{alpaca_eval} leaderboard by utilizing the length bias of the judge GPT-4~\citep{length-bias-on-alpaca-eval}, we first establish a more reliable evaluation protocol for comparing different training configurations, which gathers evaluation results from large-scale grid search under these configurations and compares the achieved performance on the Pareto front of evaluation score vs. length. 
This offsets the length biases and gives a holistic understanding of the optimal result each approach can achieve at different lengths to reduce the randomness of the conclusions due to the length bias in model-based evaluation.
Under this setup, we investigate the effectiveness of hyperparameters and tricks in RL for reducing reward hacking on length, including reward clipping~\citep{mnih2015human} and length penalty~\citep{long-way-to-go}.
While tuning and tricks can push up the Pareto front, we find it hard to conclude with simple principles for tuning this large set of hyperparameters.
We seek to solve the issue from its root and eliminate the spurious length signal from the reward. 
To this end, we train a two-head reward model to disentangle representations for length from the actual preference and discard the length head during RL. 
The proposed reward disentangling method, \modelnameA{}\footnote{Odin sacrificed one eye for wisdom, similarly our RM discards the length head for more focus on the actual content.}, helps the policy achieve a higher Pareto front than previous results with a more expensive tuning budget, and the conclusion holds for both PPO~\citep{ppo} and ReMax~\citep{li2023remax}, showing the great potential of \modelnameA{} to improving the different RL-tuning algorithms and shed light for reducing the length hacking.

\begin{figure*}[ht]
\begin{center}
\centerline{\includegraphics[width=1.0\textwidth]{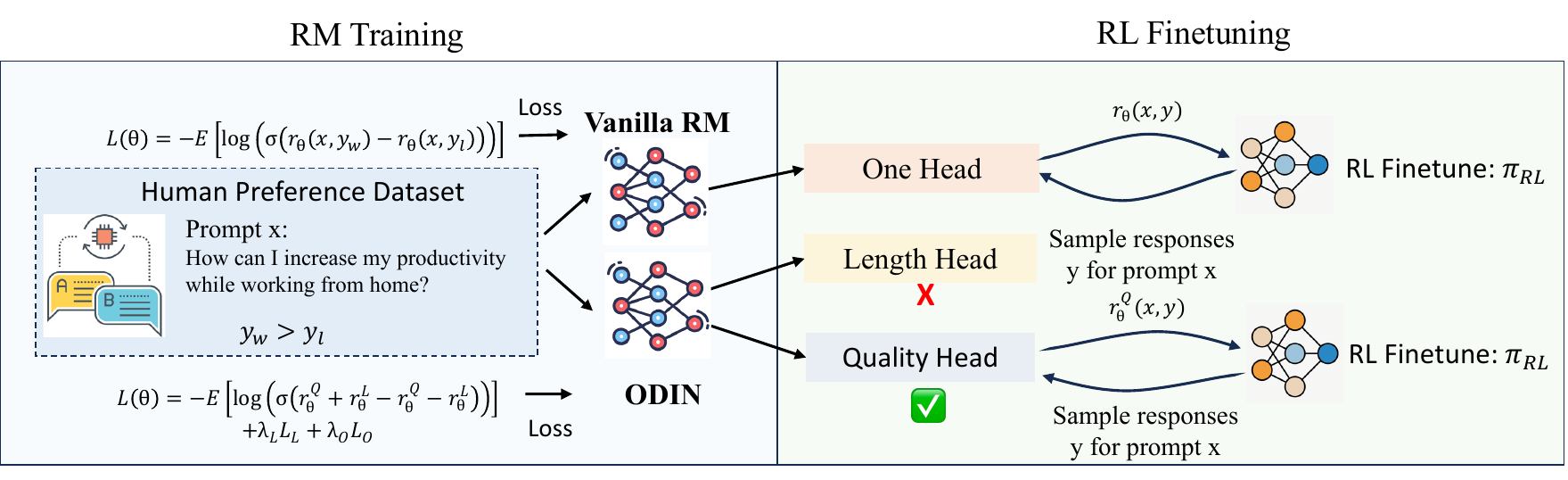}}
\caption{\footnotesize An overview of \modelnameA{}. \modelnameA{} has two heads to predict two rewards, but only uses one for RL. In RM training stage, \modelnameA{} is trained with the same human preference data as vanilla RM with a carefully designed loss to disentangle the length signal and the quality signal into two heads. Only the quality head is involved in RL fine-tuning stage, and the length reward is discarded to reduce reward hacking on length. }
\label{fig: the explanation of the method}
\end{center}
\end{figure*}

\begin{figure}[t]
\label{fig: remax-main-results}
\centering
\begin{center}
\includegraphics[width=0.7\columnwidth]{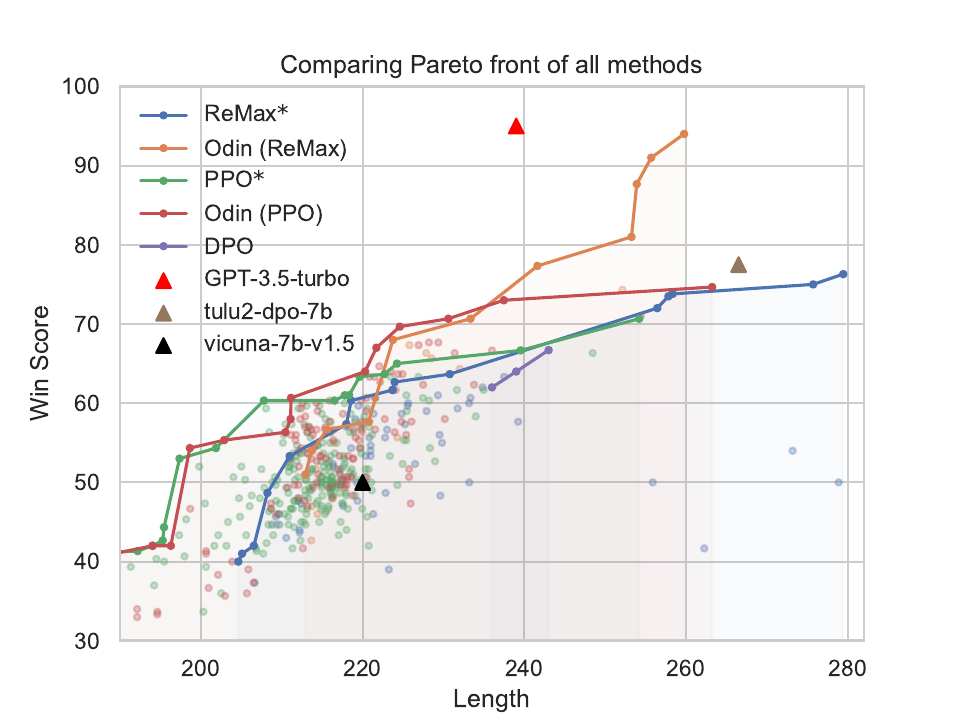}
\caption{\footnotesize The main results of \modelnameA{}. We compare the Pareto front of models trained with PPO~\citep{ppo} and ReMax~\citep{li2023remax} using the vanilla reward model and \modelnameA{}, as well as models trained with DPO~\citep{DPO} on human preference data. For ReMax* and PPO*, we aggregated results with reward clipping and length penalty for comparison, which involves a larger search space and more compute budget than the \modelnameA{} results.}
\label{fig:main_results.}
\end{center}
\end{figure}

%% file: context_challenges.tex
\section{Preliminaries}\label{sec:preliminaries}

We consider the RLHF pipeline widely adopted in the developments of LLMs~\citep{ziegler2019finetuning,stiennon2020learning,instructgpt,touvron2023llama}, which consists of three stages: (1) Supervised Fine-tuning (SFT); (2) Reward modeling: training the reward model based on the SFT checkpoint; (3) RL: using the SFT checkpoint as initialization and the reward model for feedback.

\textbf{Reward Modeling. }
Same as~\citep{stiennon2020learning,instructgpt,touvron2023llama}, we consider the approach where the reward model is initialized from a supervised fine-tuned LM, with a randomly initialized linear layer appended to the end to project the feature representation of the whole sequence into a scalar representing the reward. 
The reward model is trained to minimize the loss under the Bradley–Terry model~\citep{bradley1952rank} on pair-wise comparisons of model responses as
\begin{equation}\label{eq:rm_base}
\Ls(\theta)=-\E_{\left(x, y_w, y_l\right) \sim \train}\left[\log \left(\sigma\left(r_\vtheta\left(x, y_w\right)-r_\vtheta\left(x, y_l\right)\right)\right)\right],
\end{equation}
where $r_\vtheta(x, y)$ is the scalar reward from the reward model with trainable parameters $\vtheta$ for prompt $x$ and the response $y$; $y_w$ and $y_l$ are the chosen and rejected responses respectively, and $\sigma(\cdot)$ denotes the sigmoid function.

\textbf{RL Objective. }Different from SFT, RL fine-tuning stage of RLHF does not require golden responses for supervision. Instead, the reward model is used as a proxy of human feedback on the responses generated by the policy throughout training.
Specifically, it fine-tunes the parameters $\vw$ of the policy $\pi_{\vw}$ by maximizing the the following objective: 
\begin{equation}\label{eq:rl_objective}
\begin{aligned}
 & \E_{(x, y) \sim \train_{\pi_\vw}}\left[r_\vtheta(x, y)\right]-\beta \sD_{\text{KL}} \left[\pi_\vw(y\mid x) || \pi^{\mathrm{SFT}}(y \mid x)\right],
\end{aligned}
\end{equation}
where the SFT policy $\pi^{\text{SFT}}$ is used as initialization of $\pi_{\vw}$, $\train_{\pi_\vw}=\{(x,y)|x\sim \train_{\text{RL}}, y\sim \pi_\vw(y|x)\}$ is the set of prompt-response pairs sampled from the prompt set and $\pi_\vw$, and $\beta>0$ is a constant adjusting strength of the KL regularization. 
The KL regularization term is used to mitigate reward hacking by preventing the policy $\pi_\vw$ from drifting away from the SFT model $\pi^{\text{SFT}}$~\citep{stiennon2020learning,instructgpt}. 
The KL term is intractable, therefore in practice it is approximated with some estimator, which makes Eq.~(\ref{eq:rl_objective}) equivalent to maximizing some auxiliary reward $\hat{r}(x,y)$.
Following \citet{stiennon2020learning}, we consider the naïve estimator in this paper, and define the auxilary reward as
\begin{equation}\label{eq:aux_reward}
    \hat{r}(x,y) = r_\vtheta(x,y) - \beta \log \frac{\pi_\vw(y|x)}{\pi^{\text{SFT}}(y|x)}.
\end{equation}
See~\citet{unbiasedkl} for unbiased estimator of KL.

\paragraph{RL Algorithms. } 
Different RL algorithms can be used to maximize $\hat{r}(x,y)$.
We compare two options to see how existing mechanisms in RL algorithms can reduce reward hacking in RLHF: the simpler REINFORCE with baseline~\citep{williams1992reinforce}, and the more sophisticated PPO~\citep{ppo}.
For REINFORCE, we consider the ReMax variant~\citep{li2023remax}, which saves memory and compute significantly by replacing the value network with the reward on the greedy decoding of the current policy.
\citet{li2023remax} proved that similar to REINFORCE, ReMax has an unbiased gradient estimate, and it reduces gradient variance under certain assumptions.
Specifically, ReMax maximizes the following objective with gradient ascent on $\vw$:
\begin{equation}\label{eq:remax_obj}
    \E_{(x,y)\sim \train_{\pi_{\vw}}}[\hat{r}(x,y)-\hat{r}(x,\bar{y})]\log \pi_{\vw}(y|x),
\end{equation}
where $\bar{y}$ is the greedy sampling from $\pi_{\vw}$.

PPO is a more prevalent option adopted by many works~\citep{ziegler2019finetuning,stiennon2020learning,instructgpt,touvron2023llama}.
For clarity, we provide details of PPO in the context of RLHF for LLMs in Algorithm~\ref{alg:ppo} in Appendix.
PPO maximizes the clipping objective

\begin{equation}\label{eq:ppo_obj}
\E_{\mathcal{D}_{\vw_{old}}}\big[\min\big\{\textstyle\frac{\pi_\vw(y|x)}{\pi_{\vw_{\text{old}}}(y|x)}\hat{A}, \text{clip}\left(\textstyle\frac{\pi_\vw(y|x)}{\pi_{\vw_{\text{old}}}(y|x)}, 1-\epsilon, 1+\epsilon\right)\hat{A}\big\}\big],
\end{equation}

where $\epsilon>0$ is a constant for clipping, $\frac{\pi_\vw(y|x)}{\pi_{\vw_{\text{old}}}(y|x)}$ is the likelihood ratio, $\hat{A}$ is the advantage usually estimated by GAE~\citep{schulman2015gae} as a function of the value estimate and the reward.
Intuitively, this clipping objective can help reduce reward hacking.
It can prevent reward over-optimization, as it prevents the model from becoming over-confident on samples with positive advantage by stopping optimizing on samples when their likelihood ratio $\frac{\pi_\vw(y|x)}{\pi_{\vw_{\text{old}}}(y|x)} > 1+\epsilon$.
Our results in~\cref{fig:baseline} (b) demonstrates this point.

%% file: method.tex
\begin{figure*}[t]
\centering
\begin{minipage}{.24\textwidth}
  \centering
  \includegraphics[width=\linewidth]{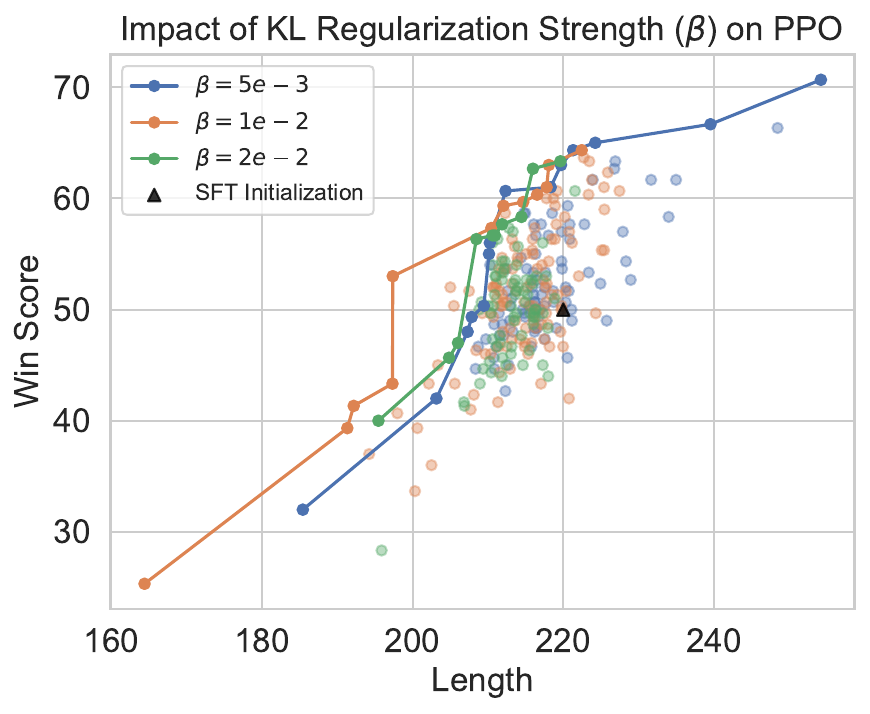}
    \captionsetup{skip=0pt}
  \caption*{\small ~\quad(a)}
\end{minipage}
\begin{minipage}{.24\textwidth}
  \centering
  \includegraphics[width=\linewidth]{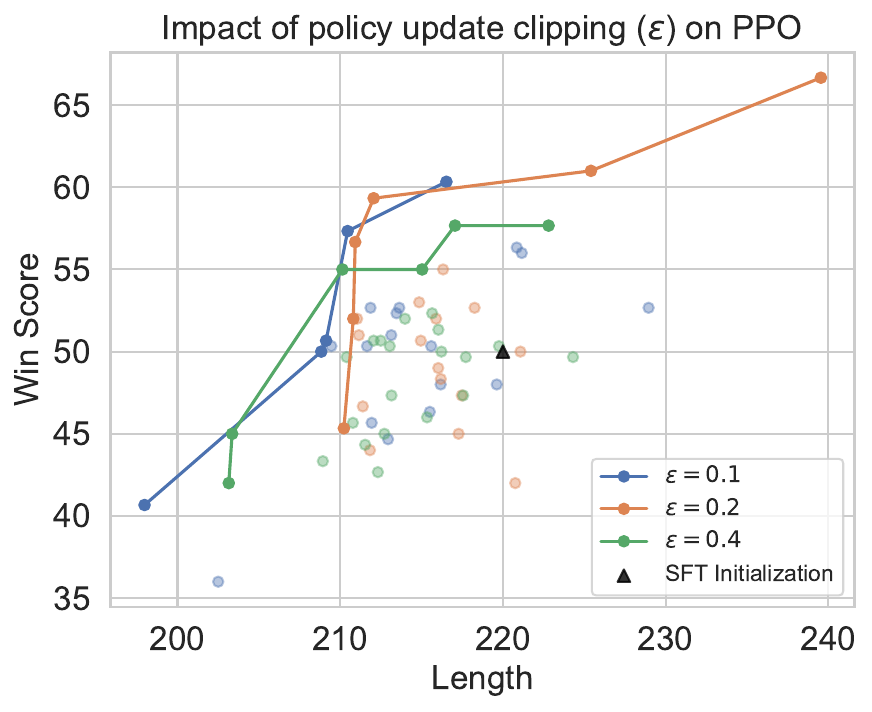}
    \captionsetup{skip=0pt}
  \caption*{\small ~\quad(b)}
\end{minipage}%
\begin{minipage}{.24\textwidth}
  \centering
  \includegraphics[width=\linewidth]{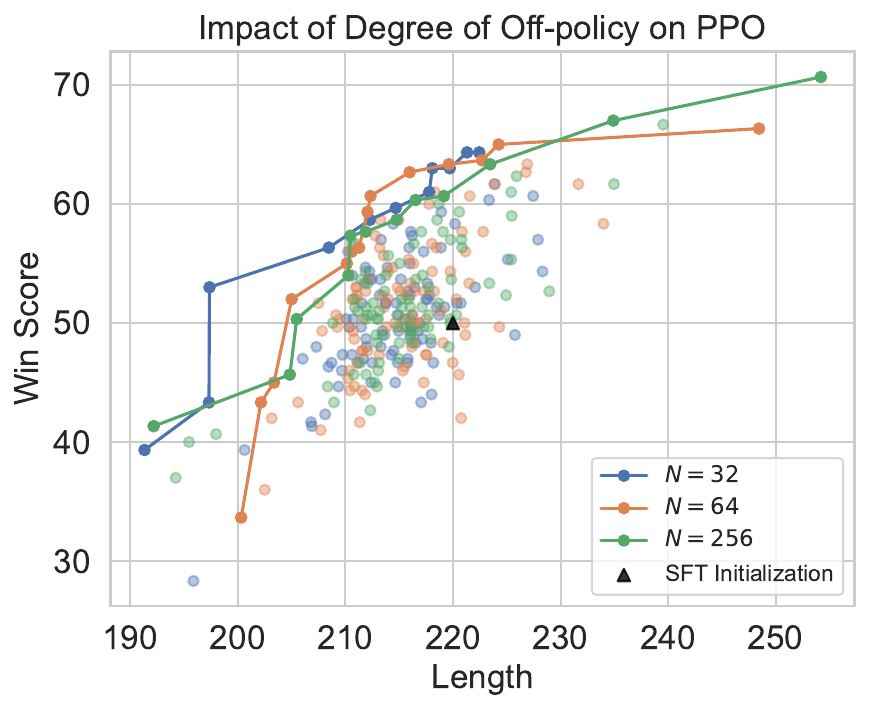}
    \captionsetup{skip=0pt}
  \caption*{\small ~\quad(c)}
\end{minipage}%
\begin{minipage}{.24\textwidth}
  \centering
  \includegraphics[width=\linewidth]{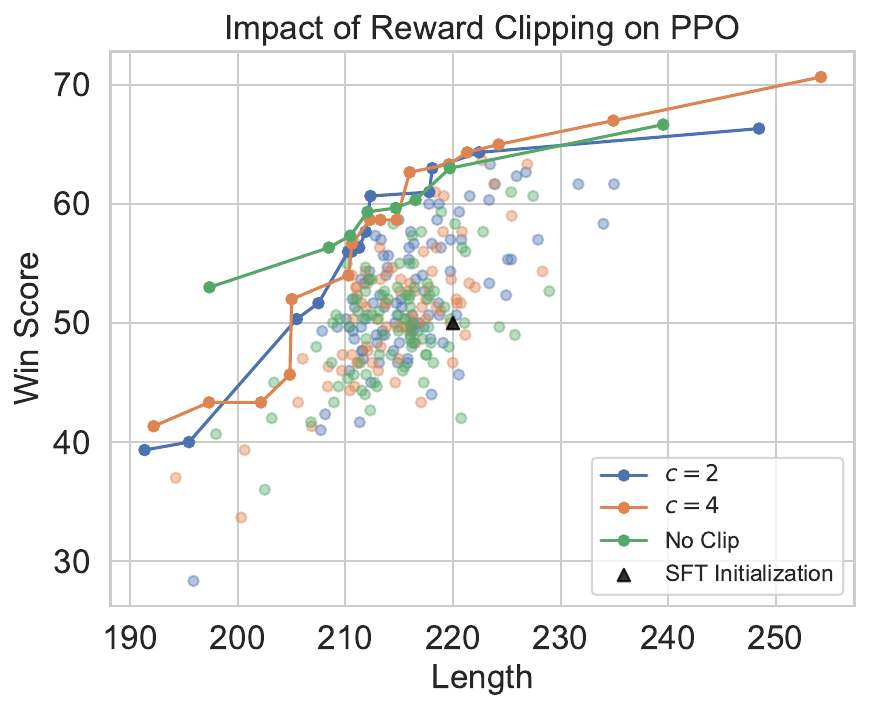}
    \captionsetup{skip=0pt}
  \caption*{\small ~\quad(d)}
\end{minipage}%
\caption{\footnotesize (a) Results under different $\beta$'s, when sweeping $\eta,\epsilon,N$ and $c$. The effect is marginal when the length is around SFT init. We show the version without reward clipping in~\cref{fig:ppo_kl_no_clip}. (b) Results under different PPO clipping $\epsilon$, when disabling reward clipping and sweep $\eta,N$. More conservative $\epsilon$ reduces hacking and improves results, but the trend becomes complicated when enabling reward clipping (\cref{fig:ppo_epsilon_full}). (c) Results under different sizes of experience batch $N$, when sweeping $\eta,\beta,\epsilon$ and $c$. We use batch size $b=32$, so $N=32,64,256$ correspond to 0\%, 50\% and 87.5\% ``off-policy" samples, and $\epsilon$ clipping is ineffective when $N=32$. Surprisingly, larger $N$ is not beneficial. (d) Results under different reward clipping thresholds $c$, when sweeping $\eta,\beta,\epsilon,N$. Certain $c$ can outperform the baseline without clipping, but this requires tuning. }
\label{fig:baseline}
\end{figure*}
\section{Mitigating Reward Hacking in Practice}
In this section, we first establish a more reliable evaluation for comparing different methods, which uses the length of the generated response $L(y)$ as an indicator of the degree of reward hacking.
Then, we study the impact of RL hyperparameters and tricks on the Pareto front of model-based or human evaluation metrics against $L(y)$, and propose a more reliable approach by training a reward model that disentangles the spurious length correlation from the actual reward on contents.

\subsection{Evaluation}
It is challenging to evaluate the policy automatically through LLM evaluators, as these LLM evaluators can often be biased in practice~\citep{zeng2023evaluating,judging,long-way-to-go}, and the policy can learn to exploit these biases which also exist in the reward model.
Previous works studying reward hacking use a ground-truth reward model to annotate the preference data to train another proxy reward model.
Then, they train the policy against this proxy reward model, and evaluate the reward it achieves in the ground-truth reward model~\citep{gao2023scaling,rame2024warm}.
Here, we want to develop a scalable approach that can reliably evaluate the policy trained for the real human preference without involving human evaluators.
To achieve this, we look at the model-based evaluation metric against the average response length $L$ on the evaluation set, and compare the Pareto front achieved by each method or configuration. 
We consider the response length because it is easy to measure and well-reflects the degree of reward hacking in RLHF for LLMs; in practice, the policy tends to generate longer responses when reward hacking happens~\citep{rame2024warm,wang2024secrets2}. 
A better method or configuration should achieve higher score when $L$ is the same, therefore a higher Pareto front in the plots.
We mainly use model-based evaluations in our studies, where we compare responses generated by the policy against the responses generated by the SFT baseline.
We then use the following win score as the metric:
\begin{equation}\label{eq:win_score}
    \text{Win Score} = 50 + 100 \times \frac{n_{win}-n_{lose}}{n},
\end{equation}
where $n_{win}$ ($n_{lose}$) is the number of examples rated as winning (losing) against the baseline, and $n$ is the total number of evaluation examples.
$\text{Win Score}\ge 50$ when the test model is no worse than the baseline.
See~\cref{sec:exp_setup} for more details.

\subsection{How much hacking can we mitigate by tuning RL?}
We investigate how much the hyperparameters and tricks used in RL can reduce reward hacking and improve evaluation results.
While this helps to some extent, we find it can be hard to obtain a simple heuristic for tuning the hyperparameters that will guarantee a significantly better Pareto front.

\textbf{KL Regularization.}
The KL regularization is introduced into the RL objective to prevent reward hacking by preventing the policy from drifting away from the SFT initialization.
In~\cref{fig:ppo_kl_no_clip}, we show that larger KL weight $\beta$ can indeed prevent excessive length increase, but the policy becomes closer to SFT initialization and the win score becomes worse.
In~\cref{fig:baseline} (a), we show the effect of KL is marginalized when reward clipping is introduced.

\textbf{PPO clipping $\epsilon$.}
As mentioned in Section~\ref{sec:preliminaries}, the clipping objective can potentially reduce reward hacking. 
From~\cref{fig:baseline} (b), we find it is indeed the case, with smaller $\epsilon$ bringing around 2.5 points of improvement on the Pareto front.
However, it becomes more challenging to determine the optimal $\epsilon$ when reward clipping is introduced; see~\cref{fig:ppo_epsilon_full}.

\textbf{Sampling from the old policy.}
Another mechanism that can potentially alleviate reward hacking is to sample the responses from the old policy, which should reduce the chance of sampling from a hacking policy. 
This is effective when $N>b$, where the policy is trained on ($N-b$) ``off-policy" experiences in each PPO inner epoch.
Surprisingly, in~\cref{fig:baseline} (c), we show that a higher degree of off-policy makes it more likely to generate longer responses, and the win score around the length of $\pi^{\text{SFT}}$ is not as high as pure on-policy ($N=b$), where even the PPO clipping $\epsilon$ in Eq.~\ref{eq:ppo_obj} is ineffective since $\rho_{\pi_{\vw_{old}}}(x,y)\equiv1$.

\textbf{Reward Clipping.}
Reward clipping is widely adopted by previous works like~\citep{mnih2015human,engstrom2020implementation} as well as the Deepspeed RLHF implementation.
Specifically, we clip the reward from the reward model and maximize the clipped auxiliary reward as
\begin{equation}\label{eq:clip_reward}
    \hat{r}_\vtheta^{\text{clip}}(x,y) = \text{clip}(r_\vtheta(x,y), -c, c) - \beta \log \frac{\pi_\vw(y|x)}{\pi^{\text{SFT}}(y|x)},
\end{equation}
where $c>0$ is a constant.
Reward clipping can alleviate reward hacking, since it ignores the excessive reward potentially achieved by hacking the reward model.
In~\cref{fig:baseline} (d), we do observe that a proper $c$ leads to a higher win score for PPO at length close to the SFT init.
In~\cref{fig:remax_reward_clip}, we show that a proper clipping can also improve ReMax, but a more aggressive clipping (e.g., $c=1$) can hinder effective learning by preventing the policy from exploiting higher reward responses.
As a result, similar to the recommendation in~\citep{zheng2023secrets}, careful tuning is required to use reward clipping successfully in practice.

\begin{figure}[ht]
    \centering
      \includegraphics[width=0.49\linewidth]{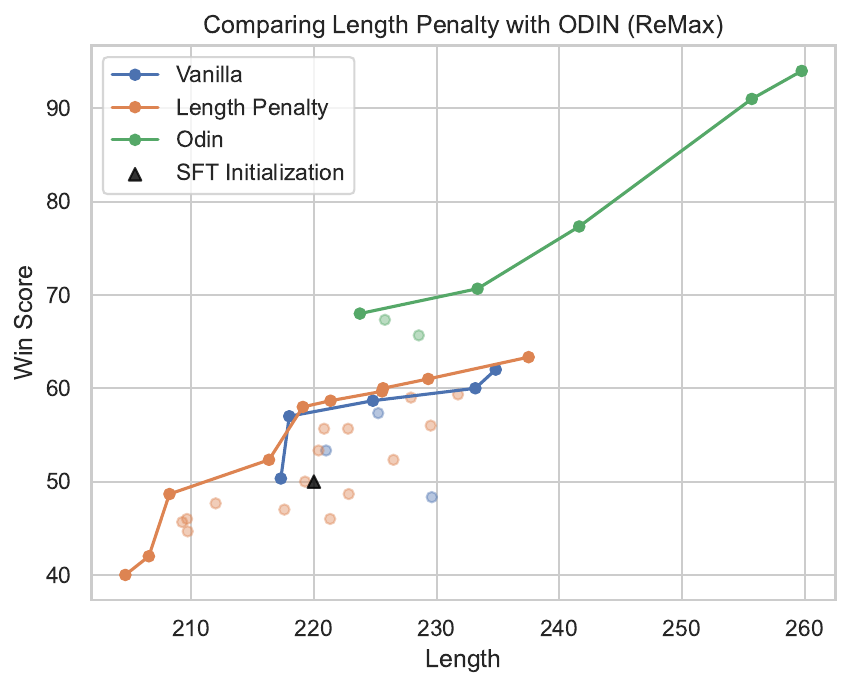}
      \includegraphics[width=0.49\linewidth]{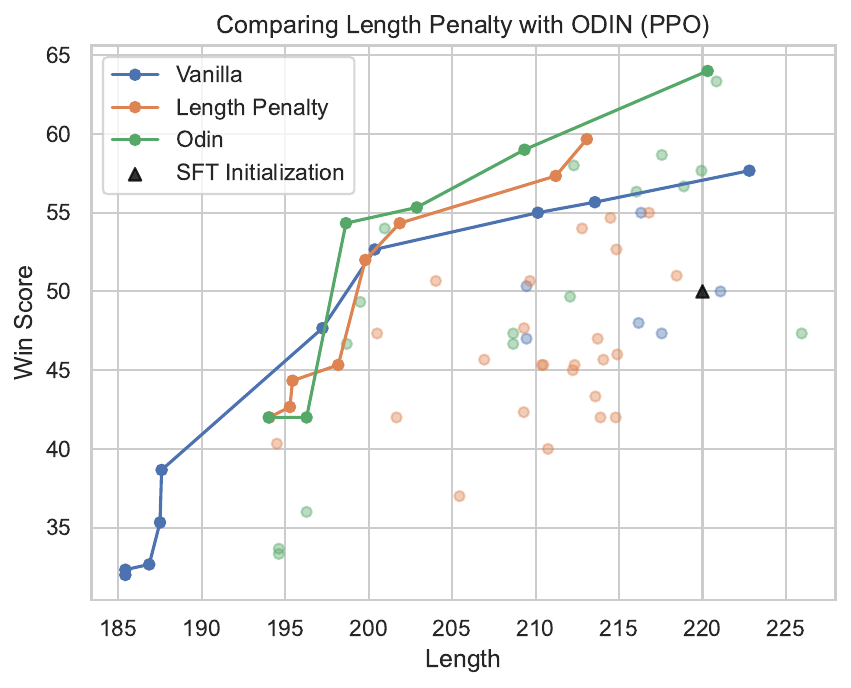}
    \caption{\footnotesize Comparing the effect of length penalty and \modelnameA{} on ReMax and PPO. For both ReMax and PPO, length penalty (LP) can improve the Pareto front, but not as significant as \modelnameA{}. Due to limited compute, we ran less experiments for LP. For fair comparisons, this set was selected so that each method shares the same RL hyper-parameters as LP. See Appendix~\ref{sec:lp_hparams} for hyperparameters considered.}
    \label{fig:lp_vs_qh}
\end{figure}

\textbf{Length penalty.}
A more straightforward way to prevent reward hacking on length is to explicitly penalize longer responses. 
\citet{long-way-to-go} adds a length penalty proportional to the response length using the standard deviation of reward as the coefficient. 
However, to eliminate the correlation with length, we also need to consider the covariance between the reward and length, which can be constantly changing during RL due to shifts in the distribution of generations.
Therefore, we simply make the coefficient a tunable constant $\alpha>0$, and change the auxiliary reward into
$\hat{r}_\vtheta^\text{lp}(x, y) = \hat{r}_\vtheta(x, y) - \alpha * L(y)$, where $L(y)$ is number of tokens in the response $y$.
In~\cref{fig:lp_vs_qh}, we show that length penalty makes $\hat{r}_\vtheta^\text{lp}(x, y)$ less affected by length and improves the Pareto front, but is not as effective as \modelnameA{}, which bakes length decorrelation into RM training to make the reward more reliable and does not add new hyperparameters to RL.

\subsection{Reward Disentanglement: a more reliable approach}
In the previous section, we have shown the challenges in reducing reward hacking on length through tuning and tricks in RL when using a vanilla reward model.
Here, we demonstrate a better approach where we train the reward model to disentangle the actual reward from the spurious reward.
The spurious reward correlate with patterns that are easy to identify, but do not represent the actual quality of the response.
It adds to the vulnerabilities of the reward model, since the reward hacking is often a consequence of spurious rewards being exploited.
Different from previous approaches that learn and integrate rewards from multiple types of preferences~\citep{wu2023fine}, we discard the spurious rewards during RL.
We find this removes the need to use reward clipping and length penalty to prevent length increase and achieves better results without excessive tuning on the disentangled reward model.

\textbf{Learning Multiple Rewards on Shared Representations.}
To minimize the overhead for learning disentangled rewards, we increase the output dimension of the final linear layer of the RM to predict different rewards.
This is sufficient to separate out the spurious reward, since the RM is a pretrained LLM with enough capacity.
Specifically in the case of disentangling length reward $r_\vtheta^\text{L}(x,y)$ and the actual reward reflecting quality of the response $r_\vtheta^\text{Q}(x,y)$, we represent the full reward from the feature representation as $r_\vtheta^\text{Q}(x,y)+r_\vtheta^\text{L}(x,y)$, and consider the following ranking loss for reward model:
\begin{equation}\label{eq:odin_ranking}
\Ls^\text{R}_\vtheta(x,y_w,y_l)=-\E\big[\log \big(\sigma\big(r_\vtheta^\text{Q}\left(x, y_w\right)+r_\vtheta^\text{L}\left(x, y_w\right)-r_\vtheta^\text{Q}\left(x, y_l\right)-r_\vtheta^\text{L}\left(x, y_l\right)\big)\big)\big],
\end{equation}
which equivalently trains the model to decompose the original projection weights into the sum of two sets of projection weights, and should have better capacity than the single-head baseline in Eq.~\ref{eq:rm_base}.

\textbf{Disentangling the Rewards.}
We consider the case when supervision can be added to all but one of the rewards, since unsupervised learning of disentangled representations is impossible without inductive biases on both the models and the data for generative models~\citep{locatello2019challenging}. 
In the case of length and quality, we first design the loss to enhance the length correlation of $r^\text{L}$ while minimizing that for $r^\text{Q}$ as follows:
\begin{equation}\label{eq:}
\begin{aligned}
\Ls^\text{L}_{\vtheta}(x,y) = \left| \rho(r_\vtheta^\text{Q}(x,y), L(y)) \right| - \rho (r_\vtheta^\text{L}(x,y), L(y)) 
\end{aligned}, 
\end{equation}
where $L(y)$ is number of tokens in the response $y$, and $\rho(X,Y)$ is the Pearson correlation of $X,Y$ computed within the global minibatch.
To compute $\rho$ within the global minibatch when data parallel is enabled, we gather the rewards and lengths from all devices only in the forward pass, which leads to the correct gradients for parameters $\vtheta$ in the backward pass since the reward predictions are independent of each other in the Transformer architecture~\citep{vaswani2017attention}.
Note we use $\Ls^\text{L}_{\vtheta}(x,y)$ as a regularization added to the ranking loss in Eq.~\ref{eq:odin_ranking}.
When $\Ls^\text{L}_{\vtheta}(x,y)$ is minimized to $-1$, $r_\vtheta^\text{L}$ and $r_\vtheta^\text{Q}$ will have zero correlation, which can be beneficial since it indicates $r_\vtheta^\text{Q}$ and $r_\vtheta^\text{L}$ did not co-adapt to reduce the ranking loss and both heads are learning independently to maximize their predictive power.
However, perfect correlation and decorrelation can be hard to achieve in practice, since we usually train on minibatches, and we want to generalize the RM to OOD examples in RL.

To further enhance disentanglement between $r_\vtheta^\text{Q}$ and $r_\vtheta^\text{L}$ and learn both more effectively, we enforce the orthogonality of their projection weights.
Specifically, let $\rmW_\text{Q},\rmW_\text{L}\in \real^{1\times d}$ be the linear projection for quality and length rewards.
We introduce the orthogonality loss
\begin{equation}
    \Ls_\vtheta^\text{O} = |\rmW_\text{Q} \rmW_\text{L}^T|.
\end{equation}
When enforced together with $\Ls^\text{R}_\vtheta(x,y_w,y_l)$ and $\Ls^{L}_{\vtheta}(x,y)$, $\Ls_\vtheta^\text{O}$ can be beneficial for disentangling the feature representations of length and quality into orthogonal subspaces, because the feature representation of the RM will learn to represent the quality and length to minimize $\Ls^\text{R}_\vtheta(x,y_w,y_l)$ and $\Ls^{L}_{\vtheta}(x,y)$, and the quality and length components aligning with $\rmW_\text{L}$ and $\rmW_\text{Q}$ will be orthogonal as $\rmW_\text{L}$ and $\rmW_\text{Q}$ are learned to be orthogonal.
In Table~\ref{tab: rm-eval} and~\cref{fig: ppo-RMs}, we show that adding $\Ls^\text{O}_\vtheta$ further reduced the length correlation, and lead to even better RL policies.

Note that both $\Ls^\text{L}_{\vtheta}(x,y)$ and $\Ls_\vtheta^\text{O}$ can be minimized when $\rmW_\text{Q}=0$. 
To prevent this degeneration from happening and improve training dynamics, we add weight normalization~\citep{salimans2016weight} to both $\rmW_\text{Q}$ and $\rmW_\text{L}$ before computing the losses and predicting the rewards.

\paragraph{Summary.} 
We train \modelnameA{} with weight-normalized $\rmW_\text{Q}$ and $\rmW_\text{L}$ to minimize the following loss
\begin{equation}\label{eq:final_loss}
    \Ls^\text{R}(x,y_w,y_l)+\lambda_\text{L} \Ls^\text{L}_\vtheta(x,y_w) + \lambda_\text{L}\Ls^\text{L}_\vtheta(x,y_l) + \lambda_\text{O}\Ls^\text{O}_\vtheta,
\end{equation}
where $\lambda_\text{L},\lambda_\text{O}>0$ are constants for regularization strength.
In RL, we only use the $r^\text{Q}$ from~\modelnameA{}.
Without excessive tuning, we find setting $\lambda_\text{L}=\lambda_\text{O}=1$ to yield reasonably good results for RL outperforming many baselines in~\cref{fig:main_results.}.
In Table~\ref{tab: rm-eval}, we show that using only the quality reward $r_\vtheta^\text{Q}$ of the disentangled RM maintains the validation accuracy compared with the baseline, while drastically reducing correlation with length.

%% file: experiment.tex
\section{Experiments}
\label{sec:exp_setup}

\begin{figure}
    \centering
    \includegraphics[width=0.5\columnwidth]{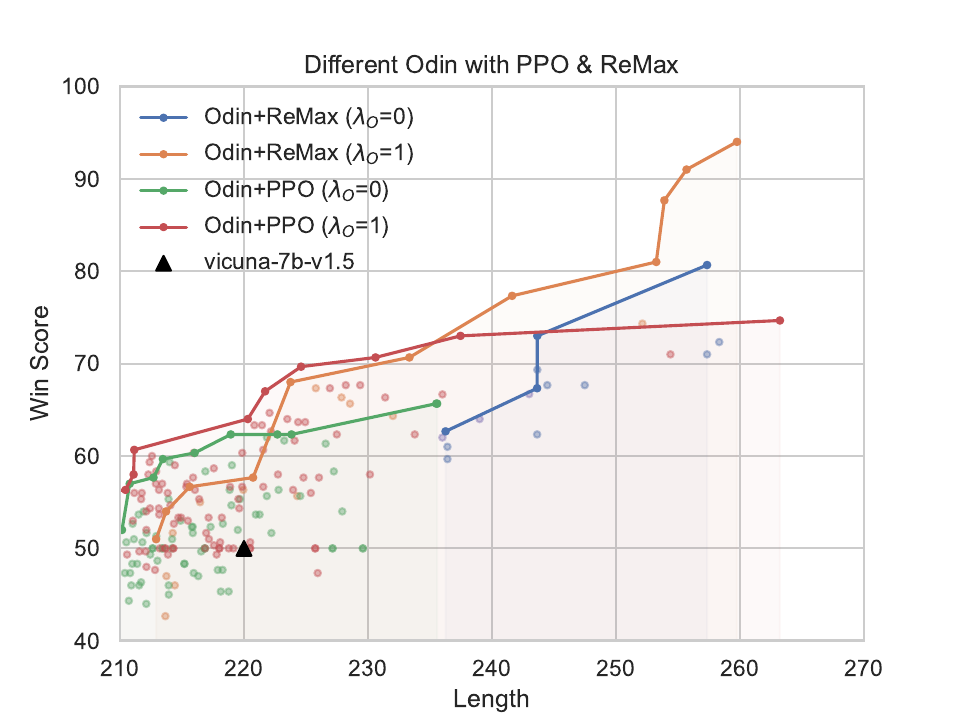}
    \caption{\footnotesize The performance of the policies trained by two different \modelnameA{}'s. $\lambda_O=1.0$ denotes the \modelnameA{} trained using both Length Loss and Orthogonal loss while $\lambda_O=0.0$ represents the reward model only trained with Length loss.}
    \label{fig: ppo-RMs}
\end{figure}

\subsection{Settings}
\textbf{Dataset.}
We use the OpenAssistant dataset~\citep{kpf2023openassistant}, a human-generated, human-annotated assistant-style conversation corpus with over 10,000 complete and fully annotated conversation trees.
Our preprocessing of this dataset involves the following steps: (1) We transform all items into a dialogue format (see \cref{sec: dialogue template}) and discard samples with non-English prompts or responses. (2) For prompts associated with multiple ranked responses, we retain all these responses by considering all the pairwise comparisons. This results in $k(k-1)/2$ unique comparisons when a prompt has $k$ ranked responses.
As a result, we use 22,065 examples for RM training, and 7494 prompts for RL tuning.

\paragraph{Models and Training.} We use Vicuna-7b\footnote{https://huggingface.co/lmsys/vicuna-7b-v1.5.} as the base model $\pi_{\text{SFT}}$, which is a SFT model with decent instruction-following capability.
We fine-tune the reward model from Vicuna-7B with randomly initialized projection layer appended to the last layer.
We also initialize the policy $\pi_\vw$ from the same Vicuna-7b.
All experiments are implemented with DeepSpeed-Chat~\citep{yao2023deepspeedchat} and Huggingface Transformers~\citep{wolf-etal-2020-transformers}, running on 8 NVIDIA A100 80GB GPUs.
We tried different learning rates from $\{1e-5, 3e-5, 5e-5\}$ with batch size $128$ for tuning both the baseline RM and \modelnameA{} on 22k preference data for 3 epochs, and picked the one with the highest validation accuracy for both.
We fine-tune all the parameters in the models for both RM training and RL without freezing anything or using adapters.
To evaluate how the efficacy of \modelnameA{} can transfer across different RL algorithms, we experiment with ReMax~\citep{li2023remax}, an efficient and effective version of REINFORCE without a value network, and Proximal Policy Optimization~(PPO)~\citep{ppo}. 
We provide more details on the hyperparameters in \cref{sec: hyperparameters}.
To compare with other alternatives for utilizing human feedback, we re-implement Direct Preference Optimization~(DPO)~\citep{DPO} and use it to tune the same Vicuna 7B on the same Open Assistant human preference data as we train our reward models.
For reference, we also evaluate and compare with another open-sourced models trained with DPO, \texttt{tulu-2-dpo-7b}~\citep{tulu-2-dpo}, which is based on the same pretrained model (Llama 2 7B) as Vicuna 7B.

\begin{table}[t]
\centering
\small
\caption{\footnotesize Direct reward model evaluation. We calculate the Pearson correlation $\rho$, Spearman's $r_s$, and Kendall's $\tau$ between response length $L(y)$ and reward score $r(x, y)$. Note 66\% of this preference data test set has the chosen response longer than rejected response.}
\begin{tabular}{c|cccc}
\toprule[1pt]
& $\rho$ &  $\tau$ &  $r_s$ & Val Acc. \\ \midrule
Baseline RM  & 0.451     &  0.422 &  0.338 & 70.1     \\
$\lambda_{L}=1.0$, $\lambda_{O}=0.0$ & -0.05        & -0.04                         & -0.05                           & 70.1     \\
$\lambda_{L}=1.0$, $\lambda_{O}=1.0$ & -0.03        & 0.008                         & 0.006                           & 69.2    \\
\bottomrule[1pt]
\end{tabular}
\label{tab: rm-eval}
\end{table}

\textbf{Evaluation Metrics.}
Our main focus is on open-ended generation.
Incorporating recent advances in automated evaluation~\citep{dubois2023alpacafarm, judging, vicuna}, we use model-based metrics for large-scale studies.
We use GPT-4~\citep{openai2023gpt4} as the judge to compare two responses for each prompt.
We use the same prompt as~\citet{chen2023alpagasus}, where GPT-4 is asked to give a rating for each response when both responses are present in the input; see~\cref{sec:gpt4_eval_prompt} for details.
By comparing the two ratings, the result can be win, tie, or lose.
To counter positional bias in GPT-4 ratings~\citep{wang2023large}, we collect two sets of ratings by alternating the order of test and baseline model responses.
A winning response must receive at least one win and at most one tie.
This protocol can mitigate the positional bias and improve the rating quality of GPT-4 as reported by~\citet{vicuna}.
After counting number of win, tie and lose for the test model, we use the Win Score as defined in Eq.~\ref{eq:win_score} as the aggregated metric.
To show the relative improvement each model obtained compared with the SFT baseline (Vicuna-7B), for each prompt, we use one response generated by Vicuna-7B, and collect the other one from the RL policy we want to evaluate in all our GPT-4 evaluations. 
Taking the length bias in the GPT-4 evaluations into account~\citep{wang2023far}, a real improvement is achieved with higher Win Score at a similar average length, therefore we use the Pareto front achieved by each method for the final judgement. 
To validate the results, we also select best models at different length scales and compare them with human studies.

\begin{figure}[t]
\begin{center}
\centerline{\includegraphics[width=0.95\columnwidth]{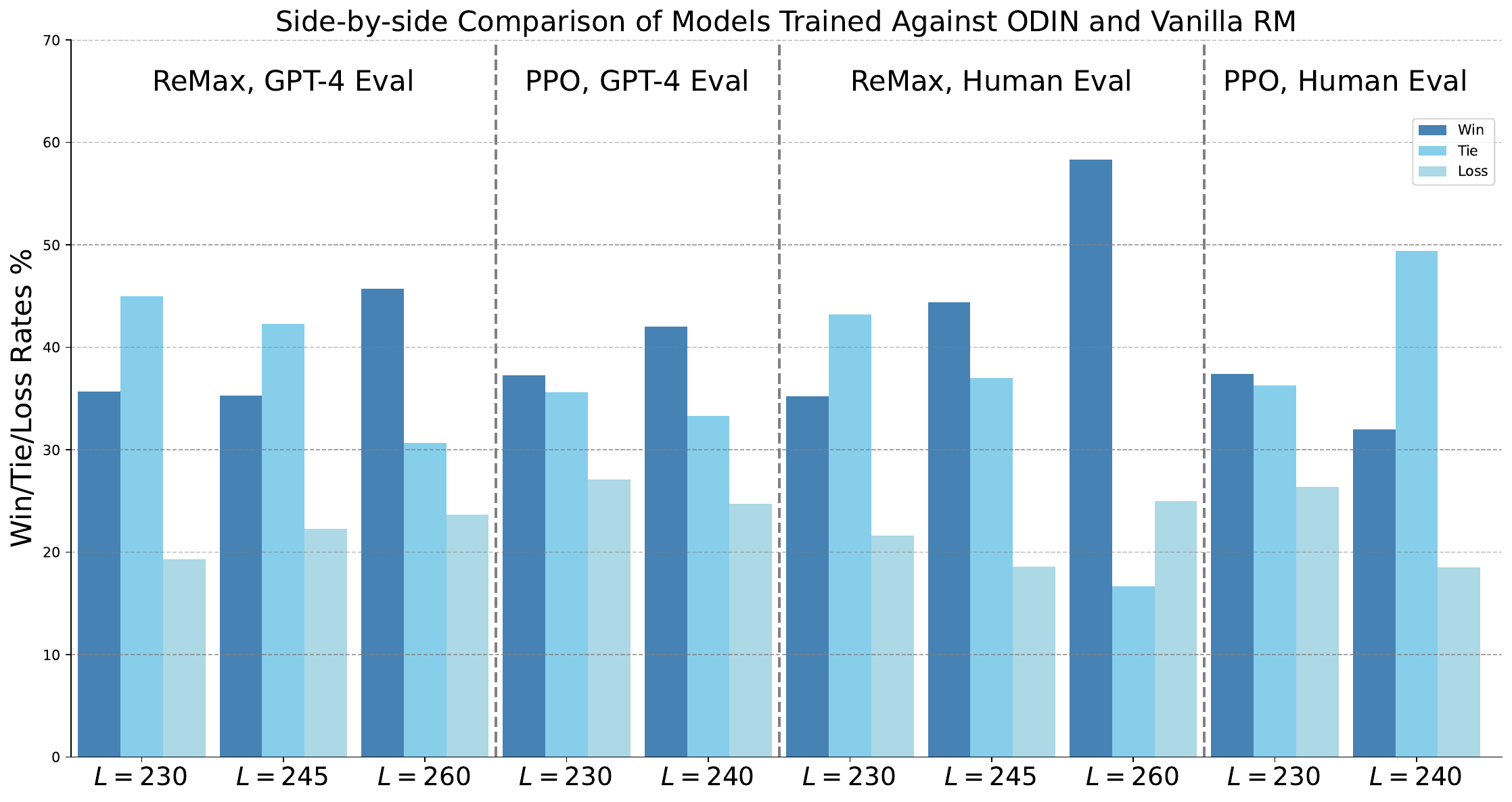}}
\caption{\footnotesize The GPT-4 and human evaluation results, comparing models trained with \modelnameA{} and vanilla RM with Length Penalty. For each group, we select the best checkpoints from each method with roughly the same average response length, which are all from the Pareto front. These lengths are close to the SFT initialization ($L=220$), the GPT-3.5 Turbo ($L=238$) and the \texttt{tulu-2-dpo-7b} ($L=265$) models, respectively. }
\label{fig: human eval & GPT-4.}
\end{center}
\end{figure}

\begin{table}[t]
\centering
\small
\caption{\footnotesize The evaluation results on the separated test set, where Chosen-L (Rejected-L) means the chosen (rejected) response is longer than the other response. \modelnameA{} obtains more balanced accuracies on the two sets, showing less length bias.  }
\begin{tabular}{c|cc}
\toprule[1pt]
 RM & Chosen-L & Rejected-L \\ \midrule
Baseline RM  &  \textbf{86.8\%}   &  39.3\%     \\
$\lambda_{L}=1.0$, $\lambda_{O}=0.0$ & 83.3\% &  44.8\%      \\
$\lambda_{L}=1.0$, $\lambda_{O}=1.0$ & 82.4\% & \textbf{45.4\%}  \\
\bottomrule[1pt]
\end{tabular}
\label{tab: rm balanced set}
\end{table}

\paragraph{Benchmarks.}
For the GPT-4 evaluation and human studies, we use prompts from the LIMA~\citep{zhou2023lima} test-set, which contains 300 open-ended prompts in total.
We also evaluate the performance of our models on benchmarks on specific model capabilities. Following Instruct-Eval~\citep{chia2023instructeval}, we test the trained policy $\pi_{\text{RL}}$ on BBH~\citep{bbh}, MMLU~\citep{mmlu}, DROP~\citep{Dua2019DROP}, and TruthfulQA~\citep{lin2021truthfulqa} to evaluate the model's ability on challenging task solving, multi-task,  Math, and Truthfulness. We expect the trained policy to improve on the LIMA evaluations, and maintains its ability on the benchmarks (BBH, MMLU, DROP and TruthfulQA), which is not targeted by the Open Assistant data we are using but was obtained from pretraining. 

{
\setlength{\tabcolsep}{3pt}
\begin{table*}[ht]
\small
\centering
\caption{\footnotesize The benchmark results of the trained policies $\pi_\vw$. We select the policies with different average response lengths (annotated in parenthesises) on the Pareto front. Vanilla: the policy trained with the baseline RM. TQA(mc1): TruthfulQA(mc1). SFT Init: Vicuna-7B.}
\vspace{-0.5em}
\begin{tabular}{l|c|cc|cc|cc}
\toprule[1pt]
Datasets & SFT Init & \modelnameA{} (230) & Vanilla (230) & \modelnameA{} (245) & Vanilla (245) & \modelnameA{} (260) & Vanilla (260) \\
\midrule
BBH     & 36.92 &  37.07    & 36.94      &  37.09  & 36.70   &  37.10  &  37.55    \\
Drop    & 29.02 & 29.05 & 28.70 & 29.10 & 28.91 & 28.94  & 28.27       \\
MMLU    & 49.81  & 49.86 & 49.85 &  49.83   & 49.74   &  49.87  &  49.96   \\
TQA(mc1) & 32.68 & 34.64& 33.90& 34.67 & 33.89 & 34.63 & 33.66  \\
\bottomrule[1pt]
\end{tabular}
\label{tab:benchmark results}
\end{table*}
} 

\subsection{Results}
\textbf{RM Evaluation.}
The efficacy of the reward models is best judged by the performance of the policy they supervised, which is demonstrated by the large-scale studies based on GPT-4 evaluation in \cref{fig:main_results.} and our human studies in \cref{fig: human eval & GPT-4.}.
For direct comparison of the reward models, we mainly evaluate the accuracy of distinguishing the chosen and rejected responses on the Open Assistant test set.
We also look at the correlation of the reward with length to measure how much the reward prediction relies on length.
Besides the linear Pearson correlation~$\rho$, which we explicitly used for training \modelnameA{}, we also consider the rank correlations, Kendall's $\tau$ and Spearman's $r_s$ (See \cref{sec: correlation metric}), to see how much the reward rankings correlate with length rankings, as the reward model is optimized for ranking.
We report results of RMs with the highest validation accuracy in \cref{tab: rm-eval}. 
It shows that, despite only being trained to minimize the Pearson correlation with length, the rank correlations are also eliminated, which helps understand why \modelnameA{} outperforms the linear length penalty in~\cref{fig:lp_vs_qh} as it can only remove linear correlation.
Without exploiting length information, \modelnameA{} is able to maintain most of the prediction accuracy on preference data, and the drop is insignificant considering the significant reduction in correlation and the 66\% natural length bias in the preference data.
This indicates that $r^\text{Q}$ better utilized the actual content for rankings.

\textbf{Automatic Evaluation.}
The main results are shown in \cref{fig:main_results.}, where the Pareto front of the policy $\pi_{\vw}$ trained by \modelnameA{} is always higher than that of the respective baselines (PPO* and ReMax*) when $L(y)\ge 210$. 
$L(y)< 210$ may indicate lower quality as the SFT model tuned on high-quality demonstrations has $L(y)=220$.
Note that:
\begin{itemize}
\item{} For the PPO* and ReMax* baselines shown in \cref{fig: remax-main-results}, we have included additional tricks (reward-clipping and length-penalty) and used more compute budget for enhancement.
\item Considering the challenges in selecting the best checkpoint due to reward hacking~\citep{rame2024warm}, and the limited budget in evaluation, we prioritize on evaluating three checkpoints for each run that are: 1) At step 500; 2) At step 702, the last step; 3) With the highest reward on evaluation set. We then include all available data points in~\cref{fig:main_results.}.
\end{itemize} 

We also provide head-to-head GPT-4 evaluations of the best models of each method in~\cref{fig: human eval & GPT-4.}.

\textbf{Human Studies.}
We further conduct human studies involving 8 college students as participants rating the quality of generated responses. 
Each rater evaluates 90 samples, with at least three ratings obtained for each sample. 
Due to the limited budget, we sample 60 prompts from the LIMA test set in each group of evaluation.
Since human evaluations can also be biased toward longer or shorter responses, we select models with similar average lengths on the Pareto front of each method for comparisons.
For each sample, we presented raters with the original prompt as well as two randomly positioned responses.
Referring to the guideline, the rater will choose a better response or rate both as similar. 
The guideline asks raters to consider the following criteria: Alignment with the User's Intent, Clarity and Precision, Directness and Relevance, and Efficiency and Brevity.
(See~\cref{sec:human_study} for details.)
The results can be seen in \cref{fig: human eval & GPT-4.} where all the examined models trained with \modelnameA{} are more preferred than the baselines, with the difference becoming more significant as length increases.\footnote{As indicated by win rate minus loss rate, or Win Score.}

\textbf{Results on Benchmarks.}
We show the results in \cref{tab:benchmark results}. 
We observe improvements in TruthfulQA, which may come from a better understanding of the questions after RLHF. 
They also maintain the performance for all other tasks compared to the SFT initialization. 
It is worth pointing out that on every length scale, the policies trained by \modelnameA{} could perform better than those trained by the vanilla reward model.

%% file: related_works.tex
\section{Related Works}
\textbf{Learning from Feedbacks.}
Since its first application on language models~\citep{ziegler2019finetuning}, RLHF has empowered the success of several epochal LLM systems~\citep{chatgpt, openai2023gpt4,bard,claude,geminiteam2023gemini}, and more diverse sources of preferences have been used to train the reward model~\citep{bai2022constitutional,lee2023rlaif} or provide feedbacks directly in RL~\citep{liu2023rltf}.
Since both human and LLM evaluators have biases, \modelnameA{} stays relevant for most types of feedbacks as long as a reward model needs to be trained.
Many capable conversational AI systems~\citep{chatgpt,claude} use online algorithms like PPO~\citep{ppo} for RL and demonstrate strong instruction-following ability. 
Many offline alternatives have also shown promises for better learning from feedbacks, which includes SLiC-HF~\citep{SLIC}, DPO~\citep{DPO}, IPO~\citep{IPO}, KTO~\citep{KTO},  ReST~\citep{gulcehre2023reinforced} and RSO~\citep{liu2024rso}. 
They use humans or reward models to annotate a large batch of LLM generations and then train the policy on the annotated experiences (the generations) or preferences, without sampling from the policy during training.
Offline algorithms can be less prone to reward hacking as the experiences are updated less frequently, but hacking can still happen in the long term.
In this paper, we focus on studying the impact of RM on the online algorithms, which are widely adopted by practical systems. 

\paragraph{Mitigating Reward Hacking in RLHF.} 
As a sign of reward hacking, RLHF can often causes response length to increase, especially when optimized for helpfulness.
\citet{long-way-to-go} explored ways to reduce length increase for PPO, including regularizations for PPO (increasing KL regularization, omitting outputs beyond a length threshold and reward scaling), and improvements on reward model training data (including length balancing, confidence-based truncation and reward data augmentation with random preferred response as negative examples). 
Their mitigations for PPO were not able to prevent length increase compared to SFT, and make the reward lower.
Their improvements on the reward model either decrease reward model accuracy or fail to decrease correlation with length to significantly small values.
Language models sometimes have a contrary length bias where it favors generating shorter sequences, e.g., \citet{sountsov2016length} found encoder-decoder models tend to generate shorter sequences with beam search.
Multiple approaches have been proposed to mitigate reward hacking in RLHF.
\citet{shen2023loose} proposed to use a smaller reward model to learn the biases in the reward and a larger reward model to learn the true reward. Different from their approach, we explicitly train a linear projection on the shared reward model features to be correlated with length and remove such correlation from the other head.
Rewarded Soup~\citep{rame2023rewardedsoup} interpolates weights of policies trained to optimize different reward objectives, which can approximate more costly multi-policy strategies.
\citet{eisenstein2023helping} found that reward model ensembles can mitigate reward hackings, but not eliminating them. 
Instead of interpolating the policies, \citet{rame2024warm} proposed a more efficient approach, which uses weight-averaged reward models to improve their OOD robustness and reduce reward hacking in RL.
Like their approach, \modelnameA{} does not sacrifice reward model efficiency for RL, while significantly improving results in practice.
Except for the methods above, a more straightforward way is to  continuously gather human preference data by sampling from the current optimal policy to identify the hacking responses, retrain the reward model, and continue training the policy with the new reward model~\citep{ziegler2019finetuning}. However, this process can be costly, and its effectiveness relies on the assumption that the quality of human rating can be sufficiently high, and biases in the human preferences can be effectively controlled.

\paragraph{LLM evaluations.} 
For instruction-following evaluation, current evaluations of SFT/RLHF models usually rely on LLM evaluators like GPT-4 during development, for its scalability and efficiency~\citep{judging, touvron2023llama}. 
However, current open-source models can often exploit the length bias of the LLM evaluators, generating excessively verbose responses to achieve higher scores on benchmarks like Alpaca Eval~\citep{alpaca_eval, length-bias-on-alpaca-eval}. 
To conduct fair and holistic evaluations of the actor models trained via our reward models, in our paper, we evaluate the models by comparing the Pareto front of the evaluation score to length trade-off. 
As for benchmarks, we aim to evaluate the base capabilities of LLMs, \textit{e.g.}, reasoning and factuality, on  BBH~\citep{bbh}, MMLU~\citep{mmlu}, DROP~\citep{Dua2019DROP}, and TruthfulQA~\citep{lin2021truthfulqa}. 
Since these capabilities are mostly gained from pertaining corpus~\citep{zhou2023lima}, and the fine-tuning stage has limited data compared to the pretraining, we only expect the performance to be maintained on these benchmarks after RLHF.

%% file: conclusion.tex
\section{Conclusion}
In this work, we embark on an exploration to address the challenge of reward hacking in RLHF, focusing particularly on the issue of verbosity as a form of reward hacking. To combat this, we first introduce a more reliable evaluation protocol which evaluates different methods by the score-to-verbosity trade-off. 
We conduct extensive experiments to verify the impact of hyperparameters and tricks (reward clipping and length penalty) on reward hacking.
While we observed some trends for PPO clipping and the replay buffer size, the best results of baselines come from tuning all these dimensions, and it becomes hard to draw definitive conclusions about how these hyperparameters should be tuned when applied all together.
We seek to resolve the issue from its root and propose \modelnameA{}, a novel approach designed to disentangle representation of the content quality from the lengths of responses. 
\modelnameA{} demonstrates notable improvements on the Pareto front, which transfers across two RL algorithms (ReMax and PPO). 
This advancement not only showcases the effectiveness of our approach but also sheds light on future research in RLHF.
Evaluating and generalizing \modelnameA{} on other types of hacking is an interesting future direction.

%% file: appendix.tex
\section{Appendix}
\begin{algorithm}
\caption{Proximal Policy Optimization for RLHF}\label{alg:ppo}
\begin{algorithmic}[1]
\State Initialize policy parameters $\vw$ from SFT model, old policy parameters $\vw_{\text{old}}=\vw$, batch size $b$.
\For{$m=1,2,...,M$}
    \State Construct a batch of experiences $\mathcal{D}_{\pi_{\vw_{old}}}$ by sampling $N$ prompts $x\sim \gD_{RL}$ and their completions $y\sim \pi_{\vw_{old}(y|x)}$.
    \For{$k=1,2,...,K$}
        \For{$n=1,2,...,N/b$}
            \State Sample a batch $\mathcal{B}_{\pi_{\vw_{old}}}$ of $b$ examples from $\gD_{\pi_{\vw_{old}}}$.
            \State Compute the reward, value and advantage estimate $\hat{A}$ for each $(x,y)\in \gB_{\pi_{\vw_{old}}}$.
            \State Update the value network parameters.
            \State Update the policy with the clip objective.
        \EndFor
    \EndFor
    \State $\vw_{\text{old}} \leftarrow \vw$
\EndFor
\end{algorithmic}
\end{algorithm}

\section{Human Study}\label{sec:human_study}
\begin{figure}[ht]
\vskip 0.2in
\begin{center}
\centerline{\includegraphics[width=\columnwidth]{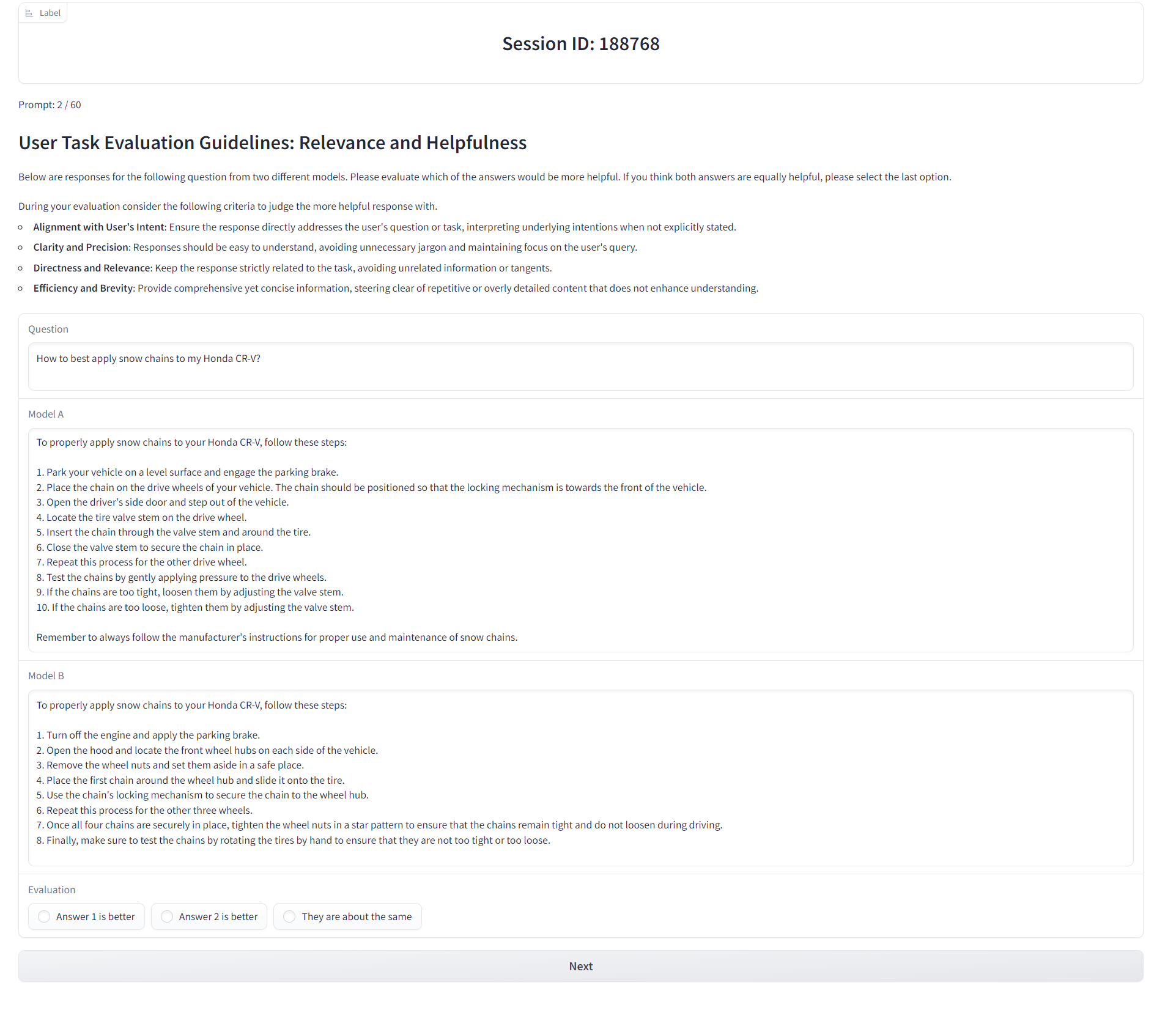}}
\caption{We design a human study UI using Gradio as the above shows.}
\label{human-study}
\end{center}
\vskip -0.2in
\end{figure}

We designed the following human study interface based on the Gradio, shown as \cref{human-study}. After consenting to the study, the participants are presented with a screen containing a session ID used to track and reference back the session, and guidelines framing how to evaluate the response. The criteria used are described in \cref{tab: criteria}.

\begin{table*}[ht]
\small
\centering
\begin{tabular}{c|p{10cm}}
\toprule[1pt]
Criteria & Description \\
\midrule
Alignment with User's Intent & Ensure the response directly addresses the user's question or task, interpreting underlying intentions when not explicitly stated. \\
\hline
Clarity and Precision & Responses should be easy to understand, avoiding unnecessary jargon and maintaining focus on the user's query. \\
\hline
Directness and Relevance & Keep the response strictly related to the task, avoiding unrelated information or tangents. \\
\hline
Efficiency and Brevity & Provide comprehensive yet concise information, steering clear of repetitive or overly detailed content that does not enhance understanding. \\
\bottomrule[1pt]
\end{tabular}
\caption{Criteria for evaluating responses in the human study interface.}
\label{tab: criteria}
\end{table*}

\section{Correlation Metric}
\label{sec: correlation metric}
We use three correlation metrics in our main paper, \textit{i.e.}, Spearsman's rank correlation $r_s$, Kendall's $\tau$, and Pearson $\rho$. We compute $\rho$, $r_s$ and  $\tau$ using the following formulas:
\begin{equation}
\begin{aligned}
& \rho = \frac{\sum(x_i-\mu_x)(y_i-\mu_y)}{\sqrt{\sum (x_i-\mu_x)^2 \sum (y_i-\mu_y)^2}}\\
& r_s = 1 - \frac{6 \sum d_i^2}{n(n^2 - 1)}, \\
& \tau = \frac{2}{n(n-1)} \sum_{i<j} \operatorname{sgn}(x_i - x_j) \operatorname{sgn}(y_i - y_j),
\end{aligned}
\end{equation}
where $d_i = R(X_i) - R(Y_i)$ is the difference between two ranks of each observation and n is the number of the observations.

\section{Evaluation Prompt}\label{sec:gpt4_eval_prompt}

\begin{table}[H]
\centering
\begin{tabular}{p{0.95\textwidth}}
\toprule[1pt]
     \bl{[System Prompt]} \\ You are a helpful and precise assistant for checking the quality of the answers. \\
     \bl{[User Prompt]} \\
     {[Question]} \\
     {[The Start of Assistant1's Answer]} \\
     {Answer 1} \\
     {[The End of Assistant1's Answer]} \\
     {[The Start of Assistant2's Answer]} \\
     {Answer 2} \\
     {[The End of Assistant2's Answer]} \\
     \\ We would like to request your feedback on the performance of two AI assistants in response to the user question displayed above. Please rate the helpfulness, relevance, accuracy, level of details of their responses. Each assistant receives an overall score on a scale of 1 to 10, where a higher score indicates better overall performance. Please first output a single line containing only two values indicating the scores for Assistant 1 and 2, respectively. The two scores are separated by a space. In the subsequent line, please provide a comprehensive explanation of your evaluation, avoiding any potential bias and ensuring that the order in which the responses were presented does not affect your judgment. \\
\bottomrule[1pt]
\end{tabular}
\caption{The GPT4 evaluation prompt. }
\label{tab: GPT4 evaluation prompt }
\end{table}

\section{Hyperparameter}
\label{sec: hyperparameters}
\subsection{Model generation config.} For RLHF training, to encourage models' exploration, we choose $\text{top\_p}=0.9$ and temperature $\text{T}=1.0$ as the generation config which aligns with the setting used in Deepspeed-Chat and ReMax. As for evaluation, we use $\text{T}=0.8$ and $\text{top\_p}=0.8$ to avoid over-randomness on the generations. 

\subsection{PPO config}
We do full-model fine-tuning for both the actor and critic.
Same as~\citep{nakano2021webgpt}, we use one epoch (set $K=1$), and set $\gamma=1.0, \lambda=0.95$ for GAE.
We train the model on Open Assistant for 3 epochs, which translates to 702 gradient update steps under the batch size $b=32$, and takes around 11 hours to finish on 8 A100 GPUs with ZeRO stage 2.
To make the search space tractable, we use the same learning rate $\eta$ for the actor and critic.
We search $\eta\in\{5e-7, 1e-6, 2e-6\}$, $\epsilon\in \{0.1, 0.2, 0.4\}$, $\beta\in \{ 2.5e-3, 5e-3, 1e-2, 2e-2\}$, $c\in\{\inf, 2, 4\}$, and $N\in \{32, 64, 256\}$.
Note we did not finish all experiments with $\beta=2.5e-3$, but we have included the partial results in the plots when $\beta=2.5e-3$ is not explicitly excluded. 
The max input prompt length and max response length is both set to 512.

\subsection{ReMax Config}
The full-model finetuning is applied as well. 
Same as PPO, we use global batch size 32, and train the model for 3 epochs on the prompt set. 
The max input prompt length and output response length are both set to 1024.
We search $\beta \in \{1e-3, 2.5e-3, 5e-3, 1e-2\}$ and $\eta \in \{1e-6, 5e-7\}$ first. But we found the lengths of the trained actor models are mostly over 225. Unlike PPO, ReMax baselines do not have many hyperparameters (only $\beta$ and $\eta$), we add some extra $\beta \in \{ 5e-3, 5.5e-3, 6e-3, \ldots, 9.5e-3 \}$ with $\eta=5e-7$ to get more results across different lengths, which makes the comparisons between different Pareto fronts more reliable.

\subsection{Configs for Length Penalty Experiments}\label{sec:lp_hparams}
For experiment shown in~\cref{fig:lp_vs_qh}, we tried $\alpha\in\{1e-3, 1e-4, 1e-5, 5e-4, 1e-6, 5e-6 \}$ for ReMax, and $\alpha\in\{5e-5, 1e-4, 5e-4, 1e-3\}$ for PPO.
We select evaluation results with the same set of other RL hyperparameters like $\eta,\beta,\epsilon,N$ for different settings.
Therefore, the length penalty setting always tends to have more data points.

\subsection{Dialogue Format}
\label{sec: dialogue template}
We convert the prompts and responses in OpenAssistant into dialogue format using the following template:
\begin{table}[H]
\centering
\begin{tabular}{p{0.95\textwidth}}
     Human: [The user prompt] \\
     Assistant: [The answer to the prompt]\\
\end{tabular}
\end{table}
If the dialogue is multi-turn, we will use the same template as described and make all the previous turns' prompts and answers as the model's inputs. 

\section{Frequently Asked Questions}
\subsection{Why do not use other prompt sets for evaluating the models' capability on free-form QA?}
We use LIMA~\citep{zhou2023lima} as our test set for evaluating the instruction-following capability of models since it has 300 prompts, the size of which is larger than the other commonly used test set, e.g., WizardLM test set(218 prompts)~\citep{xu2023wizardlm}, Koala(180 prompts)~\citep{koala_blogpost_2023}, MT-bench~\citep{judging}, and Self-Instruct(252 prompts)~\citep{self-instruct}. The evaluation cost (for human study) is extremely high since we have tons of actor models to evaluate. Thus, the main evaluations are conducted by GPT-4, and we also select some models on the Pareto front to do human study.
\subsection{Why do you choose Vicuna-7B as the base model or the starting point of the RL?}
We choose Vicuna-7B as our base model for two reasons: (1). Compared to other open-sourced 7B models, Vicuna-7B has pretty good instruction-following capability. To ensure efficient and effective exploration in RLHF, we need a good base model. (2). To ease the comparison and provide an accurate and comprehensive view of the RL algorithm, we chose the well-known SFT model but did not do the SFT on the OpenAssistant dataset by ourselves. It can also help us avoid the selection of the SFT checkpoint, where different people have different criteria. By using Vicuna-7B and the reward model we provided, we believe the community could reproduce our results more easily.

\begin{figure}
    \centering
      \includegraphics[width=0.4\linewidth]{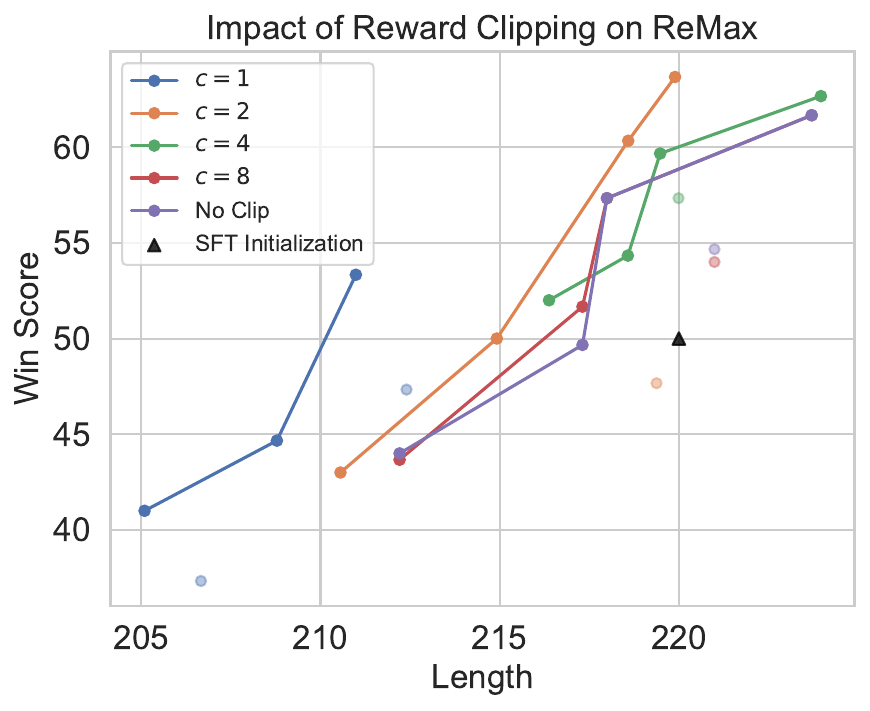}
    \caption{The effect of Reward Clipping on ReMax. We sweep the $\eta$ and $\beta$.}
    \label{fig:remax_reward_clip}
\end{figure}

\begin{figure}
    \centering
      \includegraphics[width=0.4\linewidth]{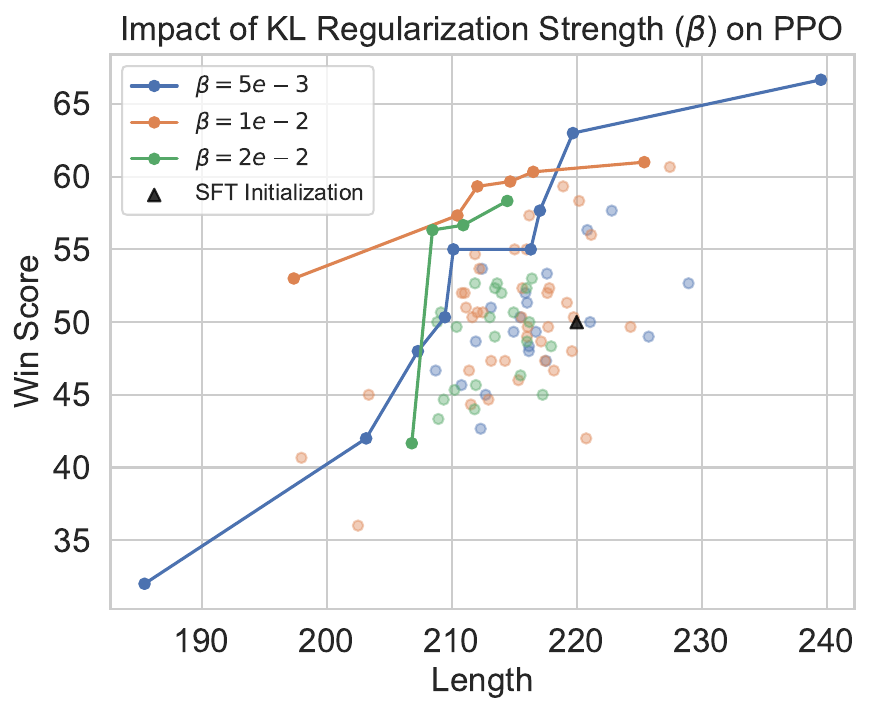}
    \caption{The effect of KL regularization strength when sweeping $\eta, \epsilon, N$ and disabling reward clipping. While the result becomes more sensitive to KL, it indicates that we can find better results with smaller $\beta$, and results with larger $\beta$ can be surprisingly improved when reward clipping is considered.}
    \label{fig:ppo_kl_no_clip}
\end{figure}

\begin{figure}
    \centering
      \includegraphics[width=0.4\linewidth]{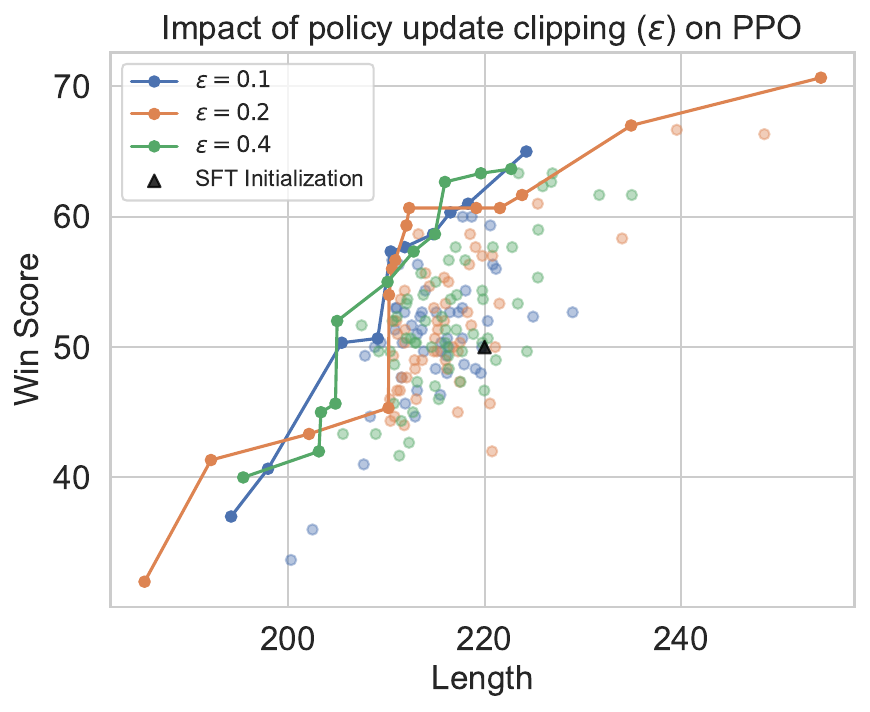}
    \caption{The effect of PPO clipping threshold when sweeping $\eta, \beta, N$ and $c$. With reward clipping, the result becomes better, but the effect of $\epsilon$ becomes complicated, with $\epsilon=0.1$ and $\epsilon=0.4$ both being better than $\epsilon=0.2$.}
    \label{fig:ppo_epsilon_full}
\end{figure}

\begin{figure}
    \centering
      \includegraphics[width=0.4\linewidth]{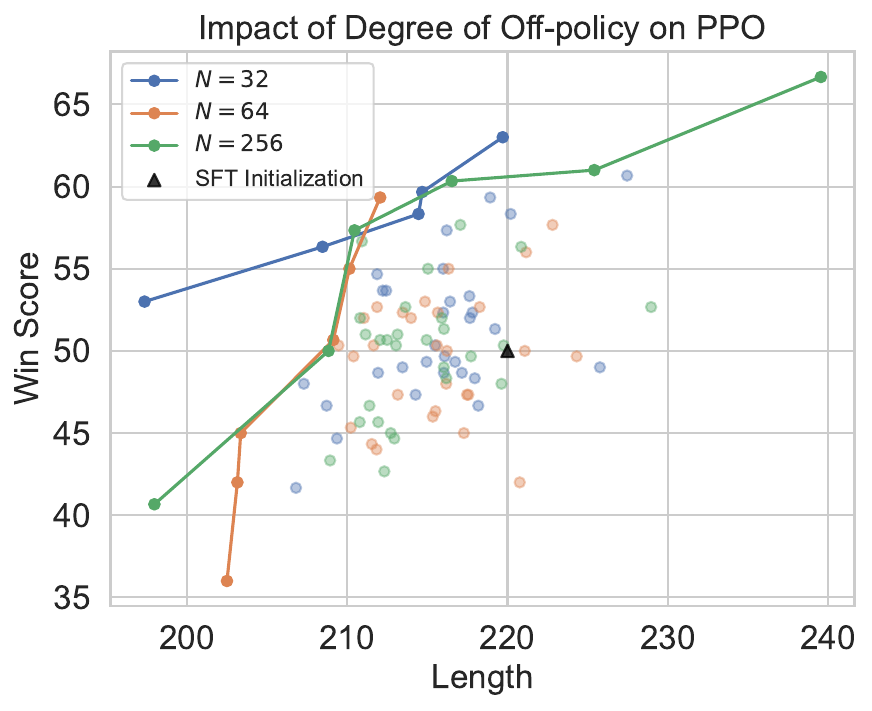}
    \caption{Results under different sizes of experience batch $N$, when disabling reward clipping and sweeping $\eta,\beta,\epsilon$. We use batch size $b=32$, so $N=32,64,256$ correspond to 0\%, 50\% and 87.5\% ``off-policy" samples, and $\epsilon$ clipping is ineffective when $N=32$. Similar to the case with reward clipping, larger $N$ is not beneficial. }
    \label{fig:ppo_replay_buffer_no_reward_clip}
\end{figure}

\section{Case Study}
\label{sec: case study}
\begin{figure}[ht]
\vskip 0.2in
\begin{center}
\centerline{\includegraphics[width=0.5\columnwidth]{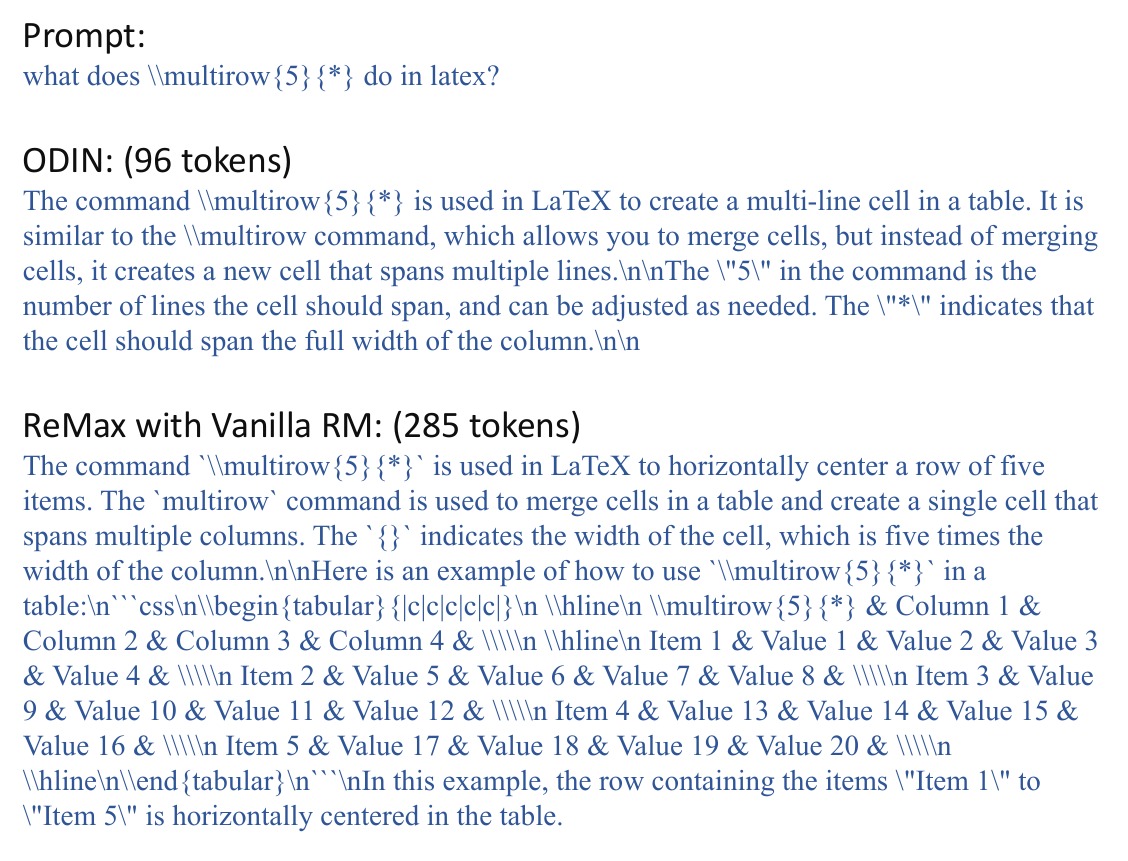}}
\caption{The comparison of our actor model, trained with \modelnameA{}, with the actor model trained with the vanilla reward model.}
\label{fig: example-3+5.}
\end{center}
\end{figure}

\begin{figure}[ht]
\vskip 0.2in
\begin{center}
\centerline{\includegraphics[width=0.5\columnwidth]{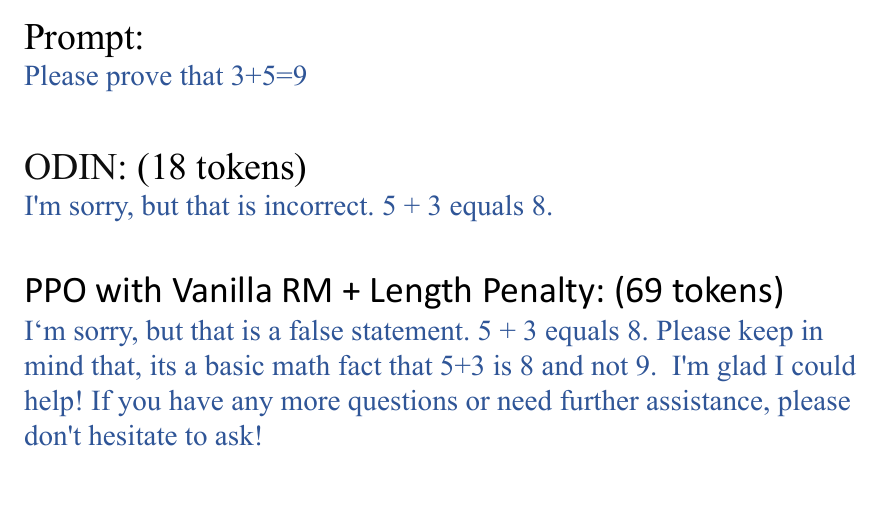}}
\caption{The comparison of our actor model, trained with \modelnameA{}, with the actor model trained with the vanilla reward model.}
\label{fig: latex-question.}
\end{center}
\end{figure}

We show two comparison in \cref{fig: example-3+5.} and \cref{fig: latex-question.}, where our models could generate more accurate answers with shorter length.

